# Enabling Human-Centered AI: A Methodological Perspective[*]

Wei Xu, SMIEEE
Center for Psychological Sciences
Zhejiang University
Hangzhou, China
xuwei11@zju.edu.cn

Zaifeng Gao
Dept. of Psychology and Behavioral Sciences
Zhejiang University
Hangzhou, China
zaifengg@zju.edu.cn

*Abstract*—**Human-centered AI (HCAI) is a design philosophy that advocates prioritizing humans in designing, developing, and deploying intelligent systems, aiming to maximize the benefits of AI to humans and avoid potential adverse effects. While HCAI continues to influence, the lack of guidance on methodology in practice makes its adoption challenging. This paper proposes a comprehensive HCAI framework based on our previous work with integrated components, including design goals, design principles, implementation approaches, interdisciplinary teams, HCAI methods, and HCAI processes. This paper also presents a "three-layer" approach to facilitate the implementation of the framework. We believe this systematic and executable framework can overcome the weaknesses in current HCAI frameworks and the challenges currently faced in practice, putting it into action to enable HCAI further.**

## I. Introduction

While a technology-centered design approach has primarily driven the design, development, and deployment of AI-based intelligent systems, the potential negative consequences of AI have mobilized researchers toward promoting a human-centered artificial intelligence (HCAI) approach [1][2]. HCAI is a design philosophy that places humans in the central place when designing, developing, and deploying intelligent systems. It aims to complement the current approaches that primarily focus on technical aspects. HCAI seeks to maximize the benefits of AI technology and address its potential adverse effects, ensuring that it serves instead of harming humans.

While the conceptual foundations of HCAI are extensively discussed in recent literature, the industry practices and methods appear to be lagging [3]. One reason is that HCAI currently consists of frameworks and design principles; there is little guidance on methodology to guide the implementation of HCAI in practice [3] [4]. Specifically, there is no comprehensive methodological framework that has been proposed to effectively guide the implementation of HCAI in practice, resulting in the challenges of adopting HCAI and considerable ambiguity about what it means to practice HCAI. People have called for HCAI methodology [3][5][6][7][8].

In the 1980s, when the computer era just started, user-centered design (UCD), as opposed to a technology-driven approach, was proposed as a design philosophy to address the user experience issues in computing systems [9][10]. Initially, UCD was presented without a mature methodological framework, but as it penetrated practices, its methodology evolved, significantly accelerating its adoption.

Similarly, as we enter the AI era, implementing HCAI requires practical methodological frameworks beyond what it can offer as a design physiology. Thus, the research question of this study is *how we can build a comprehensive HCAI methodological framework to enable its adoption in practice*.

To this end, this paper first proposes a comprehensive HCAI methodological framework based on the HCAI framework we initially proposed [1]. Then, the paper highlights the implications and challenges of implementing the proposed HCAI methodological framework. Finally, the paper offers a "three-layered" strategy for implementing this framework.

## II. Needs for a Human-Centered AI Methodological Framework

### A. Needs for a human-centered approach

The ultimate goal of AI technology is to serve humans. However, there is a double-edged sword effect for AI technology, meaning that rational use of AI will benefit humans, and unreasonable use will harm humans and society. Research has shown AI's limitations, such as vulnerability, potential bias, unexplainability, lack of causal models, development bottleneck effect, autonomy-related human factors issues, and ethical issues [1][2]. The AI Accident Database has collected over a thousand AI-related accidents [11]. The AIAAIC database shows that the number of AI abuse-related incidents has increased 26-fold since 2012 [12].

Because of these problems, some AI projects fail to use [13]. The development and use of AI technology is a decentralized global phenomenon, and the entry threshold is relatively low, which makes it more challenging and critical to control AI technology. Researchers have begun to explore a human-centered perspective in developing AI technology. Shneiderman (2020) and Xu (2019) took the lead in proposing their systematic "human-centered AI (HCAI)" conceptual frameworks [1][2]. Several other scholars have also presented their HCAI frameworks or concepts during the early exploration phase for HCAI [14][15].

HCAI aims to put human needs, values, knowledge, capabilities, and roles first during AI design, development, and deployment. The ultimate goal of HCAI is to develop human-centered AI technology (e.g., intelligent systems, tools, applications) to ensure that AI technology serves humans and enhances human capabilities rather than harming humans.

---

*B. Search for an HCAI methodological framework*

While the conceptual foundations of HCAI are extensively discussed in recent literature, the industry practices and methods appear to lag [3][4]. For example, many HCAI frameworks were defined at a high strategic level [2][14], but tactical approaches, such as HCAI-based implementation approaches, processes, and methods, were rarely integrated into these HCAI frameworks, making adopting HCAI challenging.

In the international technical community, ISO has published several standards that specify the methodology for practicing the human-centered design (HCD) approach in developing conventional non-AI-based computing systems, but there are no standards that focus on methodology to guide how to apply HCD in the development, design, and deployment of AI systems [16].

A methodological framework should include minimum components, such as design philosophy, design goals, implementation approaches, design principles, processes, and methods. Thus, some gaps impede the implementation of HCAI at the design philosophy stage; we need to develop comprehensive HCAI methodological frameworks.

### III. A Comprehensive HCAI Methodological Framework

The HCAI methodological framework presented in this paper is based on the HCAI approach that we initially proposed [1]. Xu (2019) initially proposed an HCAI approach from the perspective of the human-computer interaction (HCI) discipline. It aimed to place human needs, values, abilities, and roles at the forefront of AI design, development, and deployment and called for action from the HCI community.

Figure 1 illustrates the HCAI methodological framework. As Figure 1 shows, the HCAI framework includes the HCAI approach we presented initially, which consists of two parts (see the triangle in Figure 1): (1) the three dimensions of user, technology, and ethics, allowing us to define an HCAI approach systematically; (2) the seven primary HCAI design goals (e.g., usable AI, scalable AI, as shown in brown), allowing us to integrate specific design goals into the HCAI approach.

The HCAI methodological framework includes five components: design principles, implementation approaches, HCAI methods, HCAI processes, and interdisciplinary teams. Along with the HCAI design goals across the three dimensions of user, technology, and ethics, we present a comprehensive HCAI methodological framework to close the gaps in current HCAI frameworks. We elaborate on the components as follows.

*A. CAI design principles and design goals mapping*

Realizing HCAI design goals requires design principles that can guide project teams and organizations. Currently, most HCAI design guidelines tend to focus on strategic concepts such as human values, ethics, and privacy, which are too abstract to implement in practice, and the implementation of HCAI in system design requires more specific design guidelines [4][2][17]. Thus, clearly defining HCAI design guidelines and mapping them with the HCAI design goals will help the implementation of HCAI.

Based on existing studies and our previous research (e.g., [8][17][18][19][20], we define 28 HCAI design principles across the HCAI dimensions of user-technology-ethics; for example, "Empower users to make informed decisions and retain control." Figure 2 illustrates the mapping between the 28 design principles and the seven HCAI design goals [21].

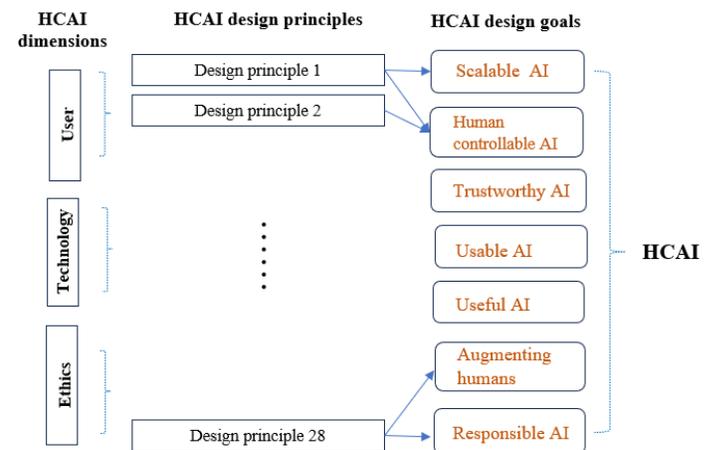

Figure 2 Mapping between HCAI design principles and goals

These design principles are specifically linked to the HCAI goals to be achieved, providing the basis for defining requirements in the early project stages of intelligent systems. Driven by HCAI design goals, applicable design principles will be chosen, helping projects clarify the HCAI direction to be followed when they define project requirements in detail. As such, these design principles provide the basis for implementing HCAI across the life cycle of intelligent systems.

*B. HCAI implementation approaches*

To promote HCAI for adoption, its methodological framework must define tactical approaches that can be implementable in the context of AI technology and current practice. Driven by the seven HCAI design goals, the framework defines 15 implementation approaches mapped explicitly to the design goals across the three dimensions of "user-technology-ethics" (see Figure 3). These implementation approaches represent how to design, develop, and deploy systems from an approach-based perspective, driving to achieve the expected design goals.

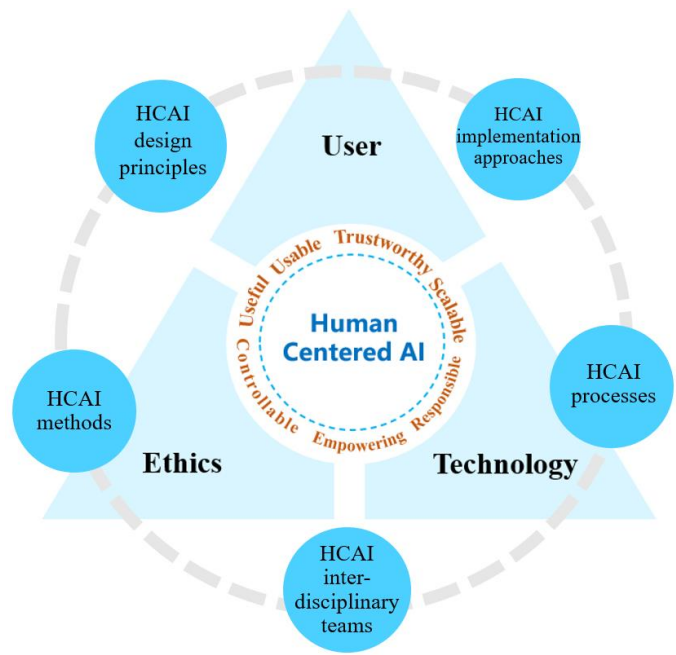

Figure 1 The HCAI Methodological Framework

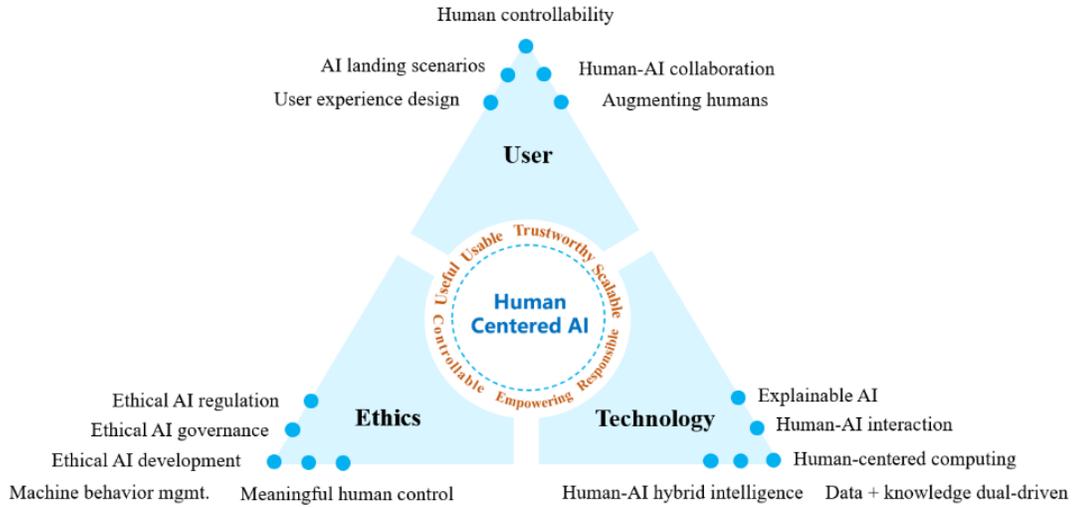

Figure 3 The HCAI framework (adapted from Xu, 2019)

Table 1 The mapping between HCAI implementation approaches, HCAI target values, and HCAI design goals

| HCAI dimensions | HCAI implementation approaches | Description of the implementation approaches | HCAI target value | Primary HCAI design goals | | | | | | |
|---|---|---|---|---|---|---|---|---|---|---|
| | | | | Useful AI | Usable AI | Augmenting humans | Trustworthy AI | Scalable AI | Responsible AI | Controllable AI |
| Technology | Human-centered computing | Integrate human values into computing (e.g., human-centered perceptual computing/social computing/algorithm nudge); Develop more powerful machine intelligence by emulating advanced human cognitive capabilities | Human value, cognitive capability | | | | | x | | |
| | Hybrid intelligence | Integrate the "human-in-the-loop"/"human-on-the-loop" mechanisms into intelligent systems for more powerful human-machine hybrid intelligence | Human role and function | | | | | x | | x |
| | "Data + knowledge" dual driven | Integrate human expert knowledge into AI through technologies such as knowledge graph, driving "data + knowledge" dual-driven AI to overcome issues (strong dependence on big data, lack of reasoning and explainability) | Human knowledge | | | | | x | | |
| | Explainable AI | Develop explainable AI algorithms and explainable UI models by following a human-centered explainable AI approach, delivering understandable AI | User trust | | x | | x | | | |
| | Intelligent interaction | Develop natural human-computer interaction technology by modeling users' physiological/cognitive/intentional/emotional states to advance AI technology | User states, experience | | x | | | x | | |
| User | UX design | Develop intelligent systems that meet user needs and are easy to use and learn through effective user experience design methods and processes | User needs, experience | x | x | | | | | |
| | AI landing scenarios | Identify valid AI landing scenarios to ensure intelligent systems are useful and solve real problems for the target users | User scenarios | x | x | | | | | |
| | Augmenting humans | Develop AI-based technologies and methods (e.g., plasticity mechanisms, controllable cognitive load) to augment human capabilities | Human capabilities | x | | x | | | | |
| | Human-AI collaboration | Develop models and methods (e.g., human-AI shared situation awareness and co-trust) to support the new design metaphor that AI can work as a collaborative teammate with humans, ensuring humans are the team leader | Human roles | x | | x | | | | |
| | Human controllability | Develop effective system designs (e.g., "human-in-the-loop", meaningful human control) to ensure that humans have the ultimate control over AI | Human controllability | | | | | | x | x |
| Ethics | Ethical AI regulations | Establish human-centered ethical AI regulations and standards to protect humans (human values, fairness, privacy, etc.) | Human values | | | | | x | x | |
| | Ethical AI governance | Implement ethical AI standards across AI lifecycle and establish a system for ethical AI review, filing, auditing, supervision, and accountability | Human accountability | | | | | | x | |
| | Ethical AI development | Enhance developers' skills in ethical AI and algorithm governance, implement ethical AI governance; track ethical AI accountability | Developer skill, behavior | | | | | | x | |
| | Machine behavior management | Adopt "human-centered machine learning," user-participatory data collection/algorithm testing/optimization" methods to avoid system output bias and unexpected behavior; effectively manage system evolving behavior | User participation | | | | | x | x | |
| | Meaningful human control | Integrate effective design mechanisms into system design (e.g., "track and trace" design for accountability) to ensure operators can make informed and conscious legal decisions for autonomous systems with sufficient information | Human accountability | | x | | | | x | |

Table 1 illustrates the relationship between the 15 implementation approaches, seven design goals, and various HCAI target values across these implementation approaches. Column 4 of the table shows these HCAI target values representing the HCAI design philosophy across many aspects such as human needs, values, abilities, roles, and so on. Thus, the execution of the 15 approaches will guarantee the implementation of placing these target values in designing, developing, and deploying AI systems, eventually achieving the "human-centered" design philosophy defined by HCAI.

These 15 implementation approaches are selected based on assessing current practices being taken or proposed across professional communities, such as AI and HCI. Many are not specific to the HCAI approach, such as explainable AI, human-AI hybrid intelligence, and ethical AI [20]. A common theme across these approaches is emphasizing one of the HCAI target values aligned with the HCAI design philosophy, providing a path to achieve the HCAI design goals. Incorporating these implementation approaches into the HCAI framework offers an integrated and systematic perspective, which maps out a "means-ends" relationship between the implementation approaches (means) and HCAI target values/design goals (ends) across the three-dimensional space of "user-technology-ethics." Thus, the methodological framework becomes executable in guiding the multidisciplinary collaboration to design, develop, and deploy HCAI systems.

### C. An HCAI-driven process

Researchers have identified the challenges of implementing HCAI in current development processes and have called for HCAI-based processes [3][6][7][8]. However, no research has presented a comprehensive HCAI process. Like an HCD process based on the HCD design philosophy, an HCAI-driven process should be across the AI life cycle based on the HCAI design philosophy, putting the human factors first in the process. Such an end-to-end process should include activities in the post-development stages, such as ethical AI governance, monitoring of AI system behavior, and so on.

To address the weakness in current HCAI practice, the HCAI methodological framework defines an integrated HCAI-driven process for implementing HCAI (see Figure 4) based on the initial concept proposed in our previous work [22]. Specifically, the integrated HCAI-driven process has three main process components, as shown in Figure 2:

- *A human-centered approach*. This is based on the "double diamond" process widely used by the HCI community, which is a human-centered process for developing HCI solutions [23]. Two diamond shapes represent four stages, emphasizing the importance of divergent and convergent thinking in problem and solution space. The "double diamond" process consists of four main stages. (1) Discovery: Understand the problem and collect user needs through user research. (2) Definitions: Analyze the collected data to identify issues that must be addressed. (3) Development: Define and validate design concepts through iterative prototyping and usability testing. (4) Delivery. Identify and build the best solution through further usability testing. Key HCAI-related activities are shown in the boxes underneath the double diamond process.
- *The AI life cycle*. This defines the typical activities to be conducted across the entire AI life cycle, including problem definition, data collection and preparation, model selection and development, model evaluation and optimization, deployment, and monitoring & maintenance. Key HCAI-related activities are shown in the boxes above the AI life cycle process.
- *HCAI guidance*. The guidance defines HCAI design goals and principles in the early stage, HCAI-based activities across design, development, and deployment stages, and HCAI-based AI governance in the later deployment stage.

Thus, the HCAI-driven process follows the HCAI design philosophy and guidance across the entire AI life cycle. It is not an abstract concept but an executable process to guide projects and address the issues in current practice [24]. Following this process will enable projects to achieve HCAI design goals.

### D. HCAI interdisciplinary teams

A fundamental driving force for advancing technology is interdisciplinary collaboration. The advance in AI technology has been benefited from collaboration with other disciplines. Researchers have called for a multidisciplinary approach for best practicing HCAI to address these challenges effectively [20]. Multidisciplinary teams are required to implement HCAI, laying out the foundation for interdisciplinary methods and processes to implement HCAI successfully.

### E. HCAI-driven methods

Research also shows a lack of effective HCAI methods in current HCAI practice. For example, non-AI professionals found it challenging to ideate many possible new interactions [3]. The HCI community has realized the need to enhance existing methods [22][25].

Wile HCAI methods are being explored, some methods are being proposed. Among these, many come from non-AI disciplines; some are enhanced methods of the existing AI disciplinary methods based on HCAI, such as human-centered explainable AI, human-centered machine learning, human-centered algorithm nudge, human-centered intelligent recommender systems, and other methods [15][26]. These methods complement existing AI disciplinary methods and help develop HCAI intelligent systems [24]. Deploying these methods will guide projects to achieve the HCAI design goals.

## IV. DISCUSSIONS

### F. Implications of the HCAI Methodological Framework

*Executable HCAI design philosophy and methodology*. The HCAI framework provides a comprehensive human-centered approach to designing, developing, and deploying AI. It addressed the weaknesses of existing HCAI frameworks and the challenges faced in current HCAI practice. It not only puts forward an AI design philosophy but also a methodology to design, develop, and deploy AI technology effectively. The framework can effectively supplement existing technical approaches to AI technology and enable AI projects to systematically coordinate HCAI-based activities across the entire AI life cycle.

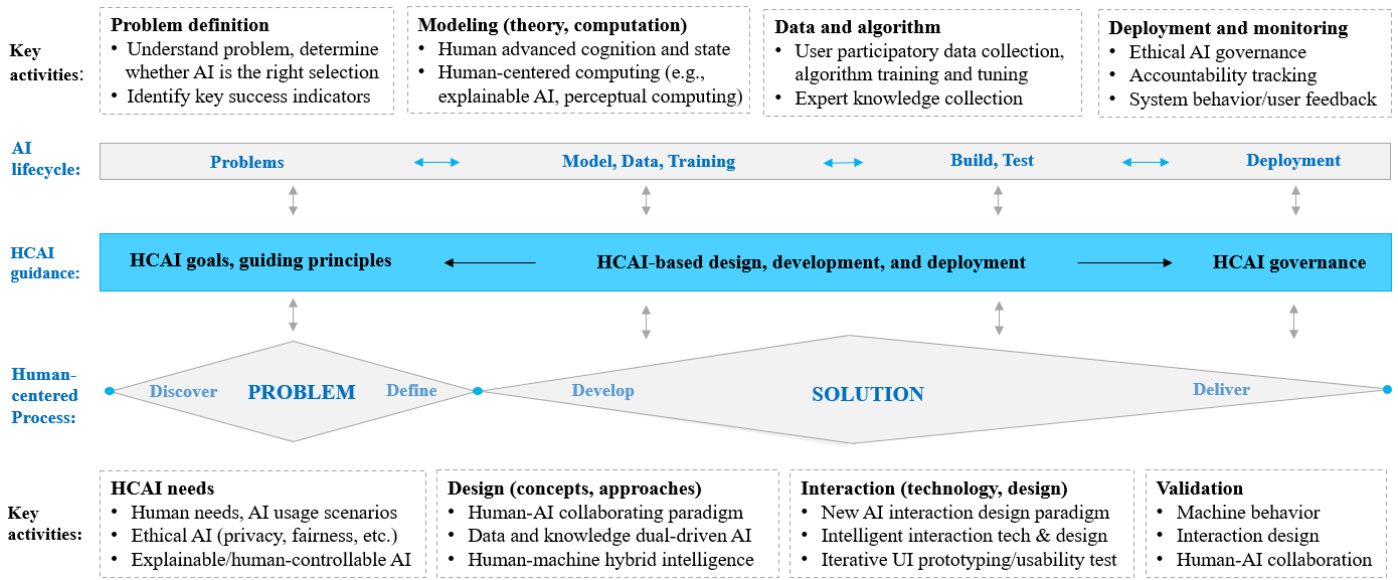

Figure 4  An HCAI process across the entire AI lifecycle

*Implementable approaches to promote HCAI.* The HCAI framework systematically puts humans (e.g., the HCAI target values such as human needs, capabilities, and functions) first in the AI design, development, and deployment across the three dimensions of " user-technology-ethics." The HCAI target values explicitly represent the HCAI design philosophy to place humans at a priority position during the execution. By mapping the seven HCAI design goals with the 15 implementation approaches in terms of these target values, the framework demonstrates the rationale of how we place humans in the first place, as claimed by HCAI. Thus, the HCAI design philosophy can be realized through the 15 implementation approaches.

*Systematic approaches.* The framework put forward systematic approaches, as represented by the following three aspects. First, based on the three dimensions of "user-technology-ethics," the framework seeks a synergy of the three perspectives from a broader human-centered perspective. Second, the framework regards humans and intelligent systems as organically integrated human-machine systems rather than siloed parts, promoting an integrated implementation approach, such as human-AI hybrid enhanced intelligence and human-AI collaboration. Third, the framework seeks to integrate interdisciplinary collaboration by promoting multidisciplinary HCAI methods, teams, and processes.

*Sustainable and scalable AI approaches.* From a long-term perspective, AI technology must be sustainable and scalable, which is one of the issues faced by the AI community. The framework seeks sustainable and scalable AI solutions from the following aspects. First, at the technical level, the framework emphasizes that AI technologies must integrate human roles by leveraging the complementary advantages of human and machine intelligence (e.g., human-AI hybrid augmented intelligence). This avoids a siloed technical development path with machine intelligence only. Second, at the user level, the frameworks fully consider various HCAI target values to ensure AI delivers value to humans. Third, at the ethical level, the framework emphasizes responsible and trustworthy AI, ensuring that AI gains the public's trust and serves humans without harming them, promoting further investment in AI technology.

*B. Recommendations for implementing the HCAI framework*

Implementing the HCAI methodological framework requires interdisciplinary collaboration. However, there is a lack of effective multidisciplinary collaboration in current practice [3]. As a result, non-AI professionals reported challenges in collaborating with AI engineers [24]. Thus, there is a need to develop a strategy that can help interdisciplinary collaboration in implementing the HCAI methodological framework.

To this end, we propose a "three-layer" HCAI implementation strategy, emphasizing the all-round promotion of HCAI practices at three levels (see Figure 5): AI project teams, AI development and deployment organizations, and a macro social environment.  Figure 5 illustrates the "three-layer" strategy across the AI life cycle. First, at the social level, we recommend the following: (1) Cultivate HCAI interdisciplinary talents as we have done for HCI talents. (2) Set up government funds to support HCAI research projects. Government AI strategies should advocate the HCAI approach. (3) Carry out cross-industry and interdisciplinary HCAI research and application projects. (4) Develop HCAI methodology-related ISO and IEEE standards, as we do today for UCD and HCI. Secondly, at the organizational level, organizations should cultivate an HCAI organizational culture, develop HCAI guidelines, and standardize processes. Lastly, at the AI project team level, we need to set up multidisciplinary teams to practice interdisciplinary methods, facilitating the execution of an HCAI process. HCAI processes will also help adopt the HCAI methodology within organizations.

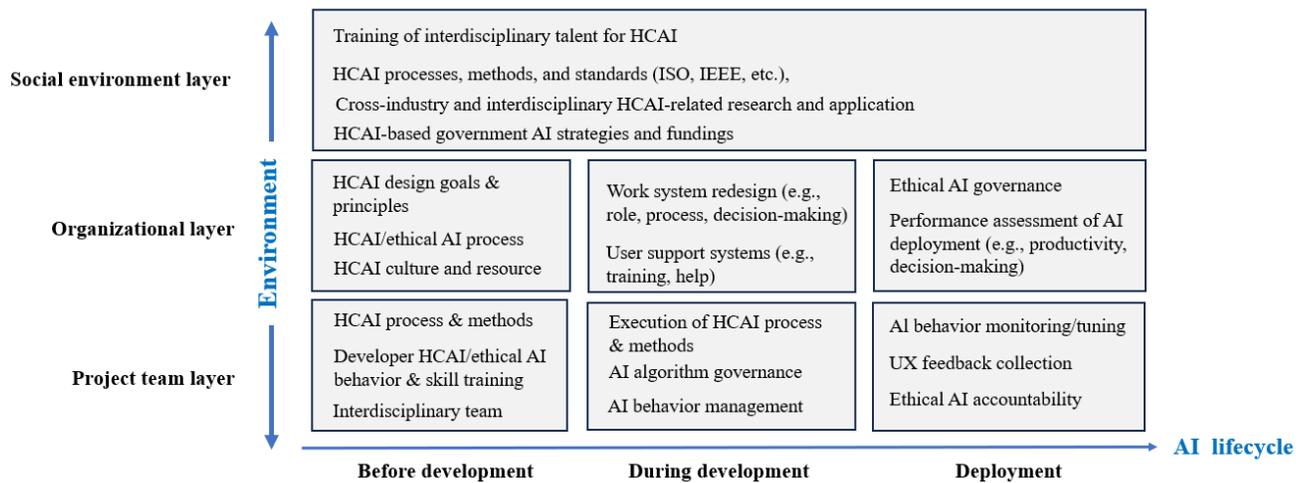

Figure 5 The "three-layer" implementation strategy

## V. Conclusions

HCAI, as a design philosophy, continues its influences on designing, developing, and deploying AI-based intelligent systems. However, the lack of a comprehensive methodological framework makes promoting and adopting HCAI challenging. This paper proposes an HCAI methodological framework to close the gap. The framework includes seven HCAI design goals, and 15 implementation approaches mapped out across the three dimensions of user-technology-ethics, presenting a systematic view of HCAI. It also includes HCAI design principles, interdisciplinary teams, HCAI-driven methods, and HCA-driven processes. To help implement the proposed HCAI methodological frameworks, this paper also proposes a "three-layer" implementation strategy.

The HCAI methodological framework makes HCAI, as a design philosophy, more executable based on methodology. The work will help overcome the weaknesses in current frameworks and the challenges currently faced in implementing HCAI, further promoting its adoption in practice. The framework can help put into action designing, developing, and deploying HCAI-based intelligent systems.